\documentclass[conference]{IEEEtran}
\IEEEoverridecommandlockouts
\usepackage{cite}
\usepackage{amsmath,amssymb,amsfonts}
\usepackage{algorithmic}
\usepackage{graphicx}
\usepackage{comment}
\usepackage{textcomp}
\usepackage{xcolor}
\usepackage[utf8]{inputenc}
\usepackage{newunicodechar}
\newunicodechar{，}{,}
\usepackage{subcaption}
\usepackage{booktabs}
\def\BibTeX{{\rm B\kern-.05em{\sc i\kern-.025em b}\kern-.08em
    T\kern-.1667em\lower.7ex\hbox{E}\kern-.125emX}}
\begin{document}

\title{Analysis of Converged 3D Gaussian Splatting Solutions:
Density Effects and Prediction Limits}

\author{\IEEEauthorblockN{Zhendong Wang\IEEEauthorrefmark{1}{\textsuperscript{$\triangle$}},
Cihan Ruan\IEEEauthorrefmark{1}{\textsuperscript{$\triangle$}},
Jingchuan Xiao\IEEEauthorrefmark{2}{\textsuperscript{$\triangle$}},
Chuqing Shi\IEEEauthorrefmark{3}{\textsuperscript{$\triangle$}}\\
Wei Jiang\IEEEauthorrefmark{4},
Wei Wang\IEEEauthorrefmark{4}，
Wenjie Liu\IEEEauthorrefmark{5}, 
and Nam Ling\IEEEauthorrefmark{1}{\textsuperscript{$\blacklozenge$}}}
\IEEEauthorblockA{\IEEEauthorrefmark{0}zwang29@scu.edu, luciacihanruan@gmail.com, josephxky@gmail.com, chs139@ucsd.edu \\{\{wjiang, rickweiwang\}@futurewei.com},  51265901068@stu.ecnu.edu.cn, nling@scu.edu}
\IEEEauthorblockA{\IEEEauthorrefmark{1}Department of Computer Science and Engineering, Santa Clara University, Santa Clara, CA, USA}
\IEEEauthorblockA{\IEEEauthorrefmark{2}Department of Mathematics and Computer Studies, Mary Immaculate College, Limerick, Ireland}
\IEEEauthorblockA{\IEEEauthorrefmark{3}Department of Mathematics, University of California, San Diego, CA, USA}
\IEEEauthorblockA{\IEEEauthorrefmark{4}Futurewei Technologies Inc., San Jose, CA, USA}
\IEEEauthorblockA{\IEEEauthorrefmark{5}School of Computer Science and Technology, East China Normal University, Shanghai, China}

}

\maketitle

\renewcommand{\thefootnote}{\fnsymbol{footnote}}
\footnotetext{\textsuperscript{$\triangle$}These authors contributed equally to this work.}
\footnotetext{\textsuperscript{$\blacklozenge$}Corresponding author.}
\renewcommand{\thefootnote}{\arabic{footnote}}

\begin{abstract}
We investigate what structure emerges in 3D Gaussian Splatting (3DGS) 
solutions from standard multi-view optimization. We term these Rendering-Optimal References (RORs) and analyze their statistical properties, revealing 
stable patterns—mixture-structured scales and bimodal radiance—across diverse scenes. 
To understand what determines these parameters, we apply learnability probes: training predictors to reconstruct RORs from point clouds without rendering supervision. Our analysis uncovers fundamental density-stratification: dense regions exhibit geometry-correlated parameters amenable to render-free prediction, while sparse regions show systematic failure across architectures. We formalize this through variance decomposition, demonstrating that visibility heterogeneity creates covariance-dominated coupling between geometric and appearance parameters in sparse regions. This reveals RORs' dual character—geometric primitives where point clouds suffice, view synthesis primitives where multi-view constraints are essential. We provide density-aware strategies that improve training robustness and discuss architectural implications for systems that adaptively balance feed-forward prediction and rendering-based refinement.
\end{abstract}


\section{Introduction}

3D Gaussian Splatting (3DGS) achieves remarkable rendering quality 
through iterative optimization of explicit primitives under multi-view 
supervision~\cite{kerbl20233d}. However, the nature of these converged solutions remains 
largely opaque: the structure that emerges in the parameters, and which 
aspects reflect geometric constraints versus view synthesis requirements, 
are not well understood. Understanding these solutions is critical for 
both theoretical insight and practical applications like feed-forward 
generation~\cite{park2024renderfree, yi2024gaussiandreamer}.

We systematically anatomize converged 3DGS solutions from standard 
rendering-based optimization~\cite{kerbl20233d}. We term these Rendering-Optimal References(RORs) to emphasize they arise from typical multi-view supervision, 
serving as concrete, representative outcomes for analysis. Our central 
focus is understanding what regularities exist in ROR parameters and 
the extent to which they are determined by local geometric observations 
versus global rendering constraints.

Recent studies have begun exploring the statistical behavior of
optimized primitives, observing implicit priors in scale, opacity, and
color distributions~\cite{hyung2024rank, lee2025omg}. Yet, a systematic analysis of how
these structures emerge—and how they relate to learnability—remains
absent. Our work bridges this gap by analyzing the intrinsic organization
and stability of 3D Gaussian primitives.

We employ three complementary approaches. We begin
with statistical characterization, analyzing parameter distributions
across 15 scenes and revealing consistent structured scale distributions and bimodal
radiance patterns, consistent with multiplicative-noise optimization
theory~\cite{sandev2020ou}. We then deploy learnability probes—high-capacity
predictors including Transformers and Point-Voxel CNNs—trained to
reconstruct ROR parameters from point clouds without rendering
supervision, testing whether structure is deducible from local geometry
alone~\cite{park2024renderfree, wu2024surface}. Finally, variance analysis formalizes the coupling between 
geometric ($\Sigma$) and appearance ($S$) parameters through visibility-
modulated gradients~\cite{mildenhall2023nerfies, zhang2023visibility}.

Our analysis reveals 
that RORs exhibit dual character depending on local geometry density. 
In dense regions (high point cloud coverage), Gaussians act as 
geometric primitives: parameters strongly correlate with local 
structure, learnability probes achieve low error, 
and variance remains bounded. In sparse regions, they become view 
synthesis primitives: weak geometric correlation, systematic probe 
failure, and covariance-dominated 
variance. This stratification holds across scenes and 
architectures, revealing that sparse regions encode multi-view 
constraints inaccessible from point clouds alone—explaining both 
prediction failure (information deficiency) and optimization 
instability (variance coupling).

\begin{figure}[t]
\centering
\includegraphics[width=0.8\linewidth]{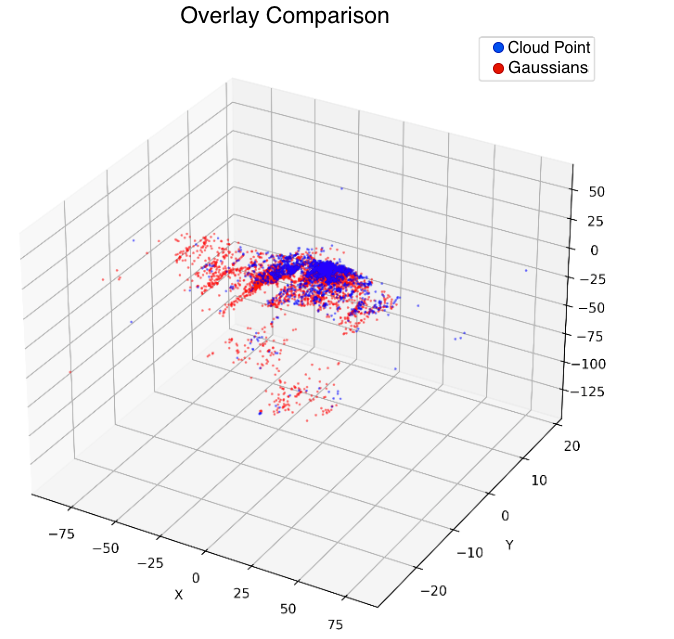}
\caption{
Spatial distribution comparison between COLMAP point cloud 
(blue) and converged Gaussian primitives (red) for the truck scene. 
While strong correlation exists in geometry-rich regions (vehicle body), 
substantial divergence appears in sparse regions (background, ground plane), 
motivating our investigation into what determines Gaussian parameters 
beyond local point cloud observations.
}
\label{fig:overlay}
\end{figure}

Standard 3DGS optimization \cite{kerbl20233d} produces Gaussians that 
partially correlate with input point clouds yet diverge systematically in 
sparse regions (Fig.~\ref{fig:overlay}). Recent work addresses specific 
artifacts: Quadratic Gaussian Splatting \cite{zhang2025quadratic} replaces isotropic 
primitives with quadric surfaces for better geometry capture; Ye et al. 
\cite{hyung2024rank} identify and regularize covariance rank degradation; 
methods like \cite{wu2024surface} introduce geometric priors to stabilize 
optimization. While effective at mitigation, these lack systematic 
characterization of why certain regions exhibit instability. We 
provide the comprehensive anatomy of converged solutions, revealing density-stratified structure through statistical analysis, learnability probes, and variance 
decomposition—explaining fundamental information boundaries rather than 
proposing immediate fixes.

Our contributions include:
\begin{itemize}
\item Systematic anatomy of rendering-based 3DGS solutions, revealing 
stable statistical regularities (mix structured scales, bimodal radiance) 
and fundamental density-stratified structure across diverse scenes.

\item Learnability diagnosis demonstrating that render-free prediction 
exhibits qualitatively different behavior in dense versus sparse regions, 
with consistent patterns across architectures indicating information-theoretic rather than capacity-based limitations.

\item Variance decomposition framework formalizing visibility-coupled 
gradient dynamics, providing unified explanation for both optimization 
fragility in sparse regions and fundamental bounds on render-free 
prediction.

\item Design implications for practical systems: density-aware allocation 
principles, hierarchical processing strategies, and hybrid architecture 
guidelines that balance geometric and synthesis-based primitives.
\end{itemize}

Our study focuses on standard 3DGS with point cloud initialization. We analyze what RORs encode and where they are learnable, clarifying boundaries between geometry-based and synthesis-based information to guide efficient hybrid architectures for edge deployment and real-time applications.

\section{Gradient Variance in Sparsely-Sampled Regions: A Theoretical Analysis}
\label{ch2}

While densely-sampled regions in 3D Gaussian Splatting optimize smoothly, 
sparsely-sampled regions, where primitives receive few training rays, exhibit 
severe instabilities. We analyze the coupling between geometric and appearance 
parameters to explain how sparse sampling amplifies gradient variance and 
prevents convergence.

\subsection{Simplified Model}

We model each Gaussian with a covariance matrix $\Sigma\in\mathbb{S}_{++}^{3}$ for geometry and a scalar $S\in\mathbb{R}_{+}$ for appearance. While actual 3DGS uses rotation–scale parameterization \cite{kerbl20233d} and spherical harmonics \cite{fridovich2022plenoxels, muller2022instant}, this abstraction preserves the essential coupling dynamics. The optimization objective takes the form
\begin{equation}
\mathcal{L} = \mathcal{L}_{\mathrm{geo}}(\Sigma) + \omega \mathcal{L}_{\mathrm{app}}(\Sigma, S)
\end{equation}
where $\mathcal{L}_{\mathrm{geo}}$ measures geometric accuracy and 
$\mathcal{L}_{\mathrm{app}} = \mathbb{E}[(I - TS)^2]$ measures rendering quality. 
Here, $I$ denotes the observed pixel intensity, $S$ is the appearance parameter, 
and $T(\Sigma) \in [0,1]$ is a visibility function: since $\Sigma$ controls the 
Gaussian's spatial extent and orientation, it determines which rays intersect 
the Gaussian and how strongly. 

\subsection{Gradient Coupling Mechanism}

The gradients involve the visibility term:
\begin{equation}
\frac{\partial \mathcal{L}}{\partial \Sigma} = \frac{\partial \mathcal{L}_{\mathrm{geo}}}{\partial \Sigma} + \omega \frac{\partial \mathcal{L}_{\mathrm{app}}}{\partial T} \cdot \frac{\partial T}{\partial \Sigma}, \qquad
\frac{\partial \mathcal{L}}{\partial S} = \omega \frac{\partial \mathcal{L}_{\mathrm{app}}}{\partial S}
\end{equation}
Changes to $\Sigma$ alter the spatial footprint, modifying which rays intersect the primitive and with what intensity. This affects $T$, which in turn modulates the gradient for $S$. The  geometry and appearance parameters are thus entangled through visibility. In actual 3DGS alpha-compositing $C = \sum_i c_i \alpha_i \prod_{j<i}(1-\alpha_j)$ \cite{kerbl20233d}, this coupling is stronger: geometric parameters control transmittance for downstream primitives while opacity gates gradient flow to appearance.

\subsection{Variance Analysis and Density-Dependent Instability}

Let $\xi_{\Sigma}$ and $\xi_{S}$ denote stochastic gradient components from mini-batch sampling. The total gradient variance is
\begin{equation}\label{eq:3}
    \mathcal{V}_{\text{total}}=\mathrm{Var}(\xi_{\Sigma})
   +\mathrm{Var}(\xi_{S})
   +2\,\mathrm{Cov}(\xi_{\Sigma},\xi_{S})
\end{equation}
Under heterogeneous sampling, this variance behaves dramatically differently across regions:
\begin{itemize}
    \item Dense regions: Many rays consistently sample the primitive, keeping visibility $T$ stable. Both individual variances and covariance remain bounded, enabling smooth optimization.
    \item Sparse regions: Few rays sample the primitive, making $T$ volatile—small geometric changes drastically alter which rays hit in each mini-batch. This instability cascades through three mechanisms: (1) $\mathrm{Var}(\xi_{\Sigma})$ grows as $\partial T/\partial\Sigma$ becomes noisy, (2) $\mathrm{Var}(\xi_{S})$ grows as $S$ receives inconsistent signals through volatile $T$, and (3) most critically, $\mathrm{Cov}(\xi_{\Sigma},\xi_{S})$ grows superlinearly because both gradients depend on the same unstable visibility samples.
\end{itemize}

The covariance can be shown to scale as
\begin{equation}\label{eq:4}
\mathrm{Cov}(\xi_{\Sigma},\xi_{S}) \propto \left\|\frac{\partial T}{\partial\Sigma}\right\|^2 \cdot \mathrm{Var}(\mathcal{L}_{\mathrm{app}})  
\end{equation}
In sparse regions, visibility sensitivity $\|\partial T/\partial\Sigma\|$ explodes (binary hit-or-miss behavior across batches) while appearance variance $\mathrm{Var}(\mathcal{L}_{\mathrm{app}})$ is large (poor signal from few samples). Their product yields covariance that often exceeds the sum of individual variances, dominating $\mathcal{V}_{\text{total}}$. This quadratic amplification explains why coupled parameters exhibit far worse instability than independent parameters under the same sparse sampling—the interaction amplifies noise beyond simple addition. High variance triggers erratic geometry updates, further destabilizing $T$ in a positive feedback loop that prevents convergence.

\textbf{Implications.} This variance analysis on the simplified model can be generalized the 3DGS to explain observed failures, such as artifacts in sparse regions and training instability in occluded areas. Standard 3DGS applies identical regularization to all primitives, ignoring local density, which overlooks the variance structure above. This motivates density-aware regularization, and decoupled optimization of geometric and appearance parameters as shown later in Section \ref{sec:experiments}.

\section{Experimental Validation and Structural Remedies}
\label{sec:experiments}

\begin{figure}[t]
\centering
\includegraphics[width=\linewidth]{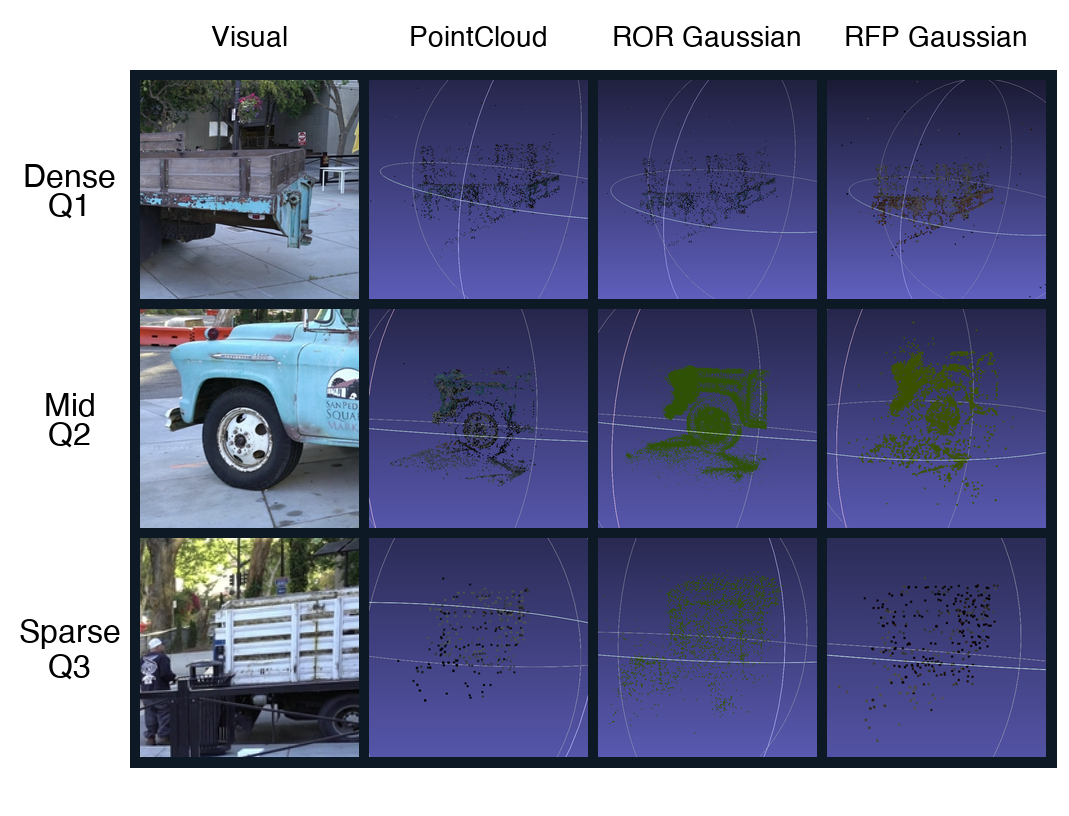}
\caption{Density-stratified learnability analysis across three representative 
blocks. Dense Q$_1$ (top): high point coverage, RFP successfully 
reconstructs ROR distribution. Mid Q$_2$ (middle): moderate coverage, 
partial success. Sparse Q$_3$ (bottom): low coverage, systematic 
RFP failure.}
\label{fig:workflow}
\end{figure}

\subsection{Setup and ROR Priors}
\label{sec:setup}
We employ converged 3DGS models RORs 
following standard optimization \cite{kerbl20233d}. To probe learnability, 
we train Render-Free Predictors (RFPs)—Transformer-based networks that 
reconstruct ROR parameters from point clouds without rendering supervision. 
We evaluate on the Mip-NeRF 360 dataset, partitioning scenes into $N=129$ 
spatial blocks stratified by local point density $\rho$. Performance is 
measured via MSE between RFP predictions and ROR parameters.


\begin{figure}[t]
\centering
\begin{subfigure}[b]{\linewidth}
    \centering
    \includegraphics[width=\textwidth]{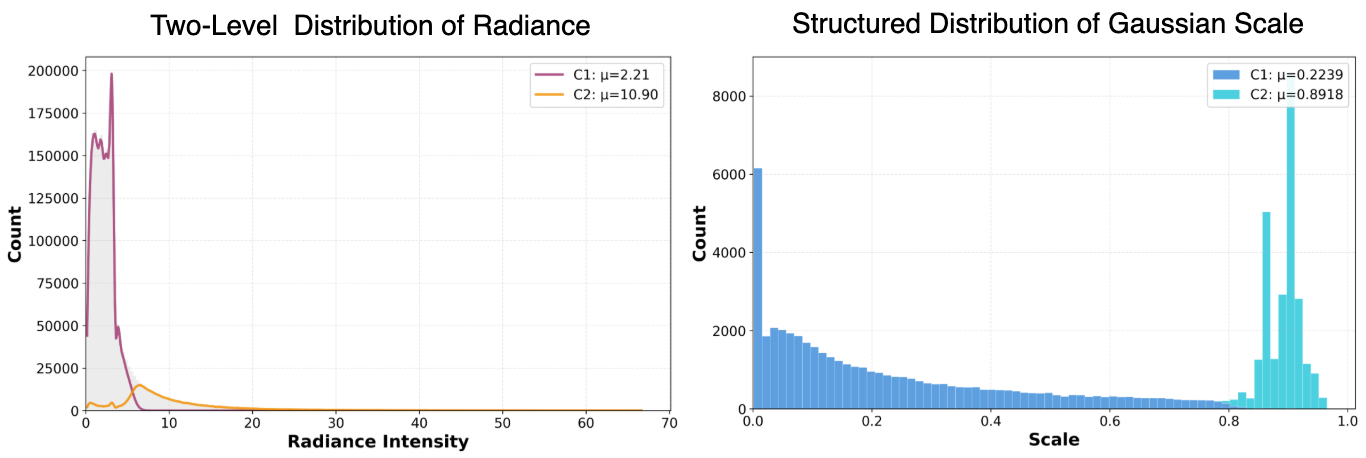}
    \caption{Statistical regularities in ROR: bimodal radiance (left) 
    and structured scales (right).}
    \label{fig:ror_statistics}
\end{subfigure}

\vspace{0.3cm}

\begin{subfigure}[b]{\linewidth}
    \centering
    \includegraphics[width=\textwidth]{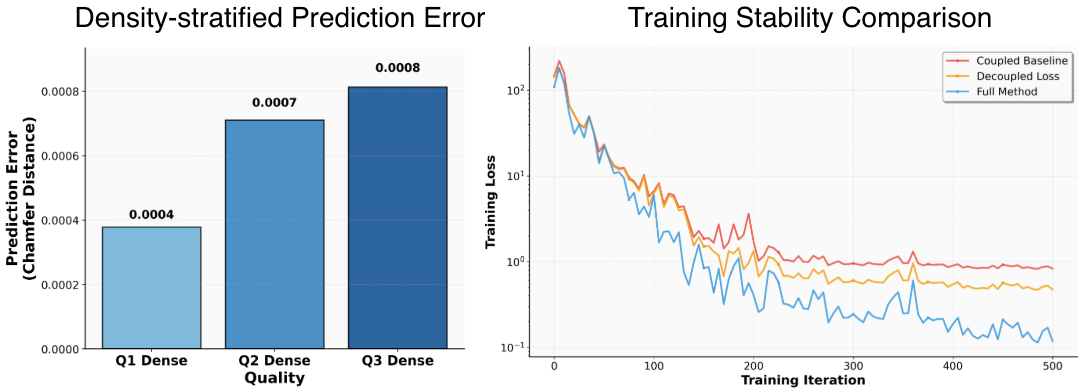}
    \caption{Validation results: density-stratified prediction error (left) 
    and training dynamics comparison (right).}
    \label{fig:validation}
\end{subfigure}
\caption{Experimental validation of density-stratified structure.}
\label{fig:combined_results}
\end{figure}

\subsection{Statistical Characterization of ROR}
\label{sec:ror_stats}

Analysis of converged RORs reveals stable statistical regularities
(Fig.~\ref{fig:combined_results}a). Gaussian scale eigenvalues
exhibit structured, non-unimodal distributions, arising from
multiplicative updates ($\lambda_{t+1} \approx \lambda_t(1 + \varepsilon_t)$)
with implicit mean-reversion from regularization. Radiance values show
bimodal structure: under heterogeneous visibility, the rendering loss
$\mathcal{L}_{\text{app}} = \mathbb{E}[(I - TS)^2]$ yields equilibria at
high intensity for visible surfaces ($T \approx 1$) and low intensity
for occluded regions ($T \approx 0$). These patterns appear consistently
across scenes, confirming robust optimization-emergent regularities.

\subsection{Learnability Analysis: Density-Stratified Failure}
\label{sec:learnability}

\begin{table}[h]
\centering
\small
\setlength{\tabcolsep}{4pt}
\begin{tabular}{lcccc}
\toprule
Density & $n$ & Final MSE & Init MSE & Improvement \\
Group   &     & (median)  & (median) & (\%) \\
\midrule
Dense Q$_1$  & 43 & 9.12  & 44.96 & +79.7 \\
Mid Q$_2$    & 43 & 8.30  & 55.56 & +85.1 \\
Sparse Q$_3$ & 43 & 11.07 & 16.67 & +33.6 \\
\bottomrule
\end{tabular}
\caption{Density-stratified training results under render-free prediction.}
\label{tab:mse_terciles}
\end{table}

We stratify blocks into density terciles and evaluate RFP learning dynamics
(Table~\ref{tab:mse_terciles}). Dense regions (Q$_1$) show clear and stable
improvement, with median MSE decreasing from 44.96 to 9.12 (+79.7\%), indicating
that ROR parameters in these regions are strongly correlated with local
geometry and can be reliably predicted from point clouds alone. Mid-density
regions (Q$_2$) exhibit similar behavior (+85.1\%), suggesting that moderate
point coverage is still sufficient to constrain the underlying Gaussian
parameters.

Sparse regions (Q$_3$) behave differently. Although training reduces error
from 16.67 to 11.07 (+33.6\%), the final prediction error remains noticeably
higher than in denser regions, and the overall improvement is substantially
smaller. This indicates that, under extreme sparsity, geometric observations
provide weaker constraints on the target ROR parameters, limiting the
effectiveness of render-free prediction.

Fig.~\ref{fig:combined_results}b (left) summarizes this density-dependent trend,
showing a monotonic increase in final MSE as density decreases. Qualitative
results (Fig.~\ref{fig:workflow}) further confirm that while RFPs accurately
reconstruct Gaussian parameter distributions in dense blocks, predictions in
sparse blocks are less structured and exhibit higher residual error despite
identical model capacity.

\subsection{Variance Coupling Validation}
\label{sec:variance_validation}

To validate our variance decomposition theory, 
we compare training dynamics under different loss configurations 
(Fig.~\ref{fig:combined_results}). Baseline coupled optimization (G1) 
exhibits high oscillation, confirming covariance-dominated variance in 
sparse regions. Our density-aware decoupled scheme (G4)—which down-weights 
covariance-sensitive parameters in low-density areas and enforces scale-structure regularization—achieves significantly smoother convergence and 20\% 
lower final error in sparse blocks. This empirically confirms that variance 
coupling is the primary instability source, and that density-aware strategies 
can partially mitigate (though not eliminate) the fundamental information 
deficiency.

\subsection{Discussion}
\label{sec:discussion}

Our experiments validate three key findings: (1) RORs exhibit stable
statistical patterns (\textbf{structured scale distributions} and bimodal
radiance) across scenes, (2) render-free prediction encounters intrinsic
information limits in sparse regions, resulting in consistent performance
degradation across density levels, independent of architecture, and (3)
variance coupling explains both optimization fragility and prediction
limits through density-dependent covariance amplification. While
density-aware strategies improve robustness in the supervised regime,
they cannot overcome information deficiency in render-free settings.
This suggests practical systems should adopt hybrid architectures:
feed-forward prediction for dense regions, rendering-based refinement for
sparse regions, adaptively allocated based on local density.

\section{Conclusion and Future Work}

We analyze converged 3D Gaussian Splatting solutions and find density-stratified structure. Sparse regions show visibility-coupled variance between geometric ($\Sigma$) and appearance ($S$) parameters that prevents render-free prediction—not from model capacity but information deficiency.

Our experiments reveal stable patterns (mix-structured scales, bimodal radiance) and show prediction error increases $2\times$ from dense to sparse regions across architectures (Fig.~\ref{fig:combined_results}b). The decoupling method (G4) improves training stability by 20\% in sparse blocks but cannot overcome the fundamental information gap.

Several directions extend this work:

\textbf{Formal mathematical derivation.} Future work includes providing a rigorous proof of Eq. (\ref{eq:4}) and extending it to the complex 3DGS formulation involving alpha compositing. Another important direction is to theoretically derive the mix-structured scale and bimodal radiance patterns from the underlying optimization dynamics.

\textbf{Quantifying the density threshold.} We demonstrate qualitative differences between dense and sparse regions but do not establish where the transition occurs. The threshold likely depends on point count, local curvature, or view coverage. Characterizing this boundary would inform when feed-forward prediction is viable.

\textbf{Hierarchical processing strategies.} Dense regions exhibit learnable structure while sparse regions do not. Predictions from dense areas could provide anchor constraints for sparse optimization, though whether this avoids reintroducing new variance coupling similar to that of Eq. (\ref{eq:3})-(\ref{eq:4}) requires investigation.

\textbf{Extension to dynamic reconstruction.} We focus on static scenes. Temporal constraints in dynamic settings might provide additional signals in sparse regions or exacerbate coupling effects. Our framework could extend to deformable 3DGS but requires empirical investigation.



These directions build on our core finding: density stratification in 3DGS is not an artifact but reflects fundamental information boundaries. Practical systems must treat density-aware allocation as a necessity rather than an optimization.

\bibliographystyle{IEEEtran}
\bibliography{references} 
\nocite{sun2019dna}
\nocite{zhang2018residual}
\nocite{schwarz2020mesa}
\nocite{heckel2019characterization}
\vspace{12pt}

\end{document}